\documentclass[sigconf]{acmart}
\usepackage{amsfonts}
\usepackage{graphicx}
\usepackage{amsmath}
\usepackage{enumerate}
\usepackage{algorithm}
\usepackage{algpseudocode}
\usepackage{color}
\usepackage{multirow}
\usepackage{booktabs, makecell, tabularx}
\usepackage{libertine}
\usepackage{colortbl}
\usepackage{textcomp}
\usepackage{tcolorbox}
\usepackage{balance}
\tcbuselibrary{listings, breakable}
\AtBeginDocument{%
	}

\copyrightyear{2025}
\acmYear{2025}
\setcopyright{acmlicensed}\acmConference[KDD '25]{Proceedings of the 31st ACM SIGKDD Conference on Knowledge Discovery and Data Mining V.2}{August 3--7, 2025}{Toronto, ON, Canada}
\acmBooktitle{Proceedings of the 31st ACM SIGKDD Conference on Knowledge Discovery and Data Mining V.2 (KDD '25), August 3--7, 2025, Toronto, ON, Canada}
\acmDOI{10.1145/3711896.3736973 }
\acmISBN{979-8-4007-1454-2/2025/08}

\begin{document}
	
	\title{FusionSAM: Visual Multi-Modal Learning with Segment Anything Model}
	
	\author{Daixun Li}
	\orcid{0009-0006-8689-6929}
	\affiliation{%
		\institution{Xidian University}
		\city{Xi'an}
		\state{Shaanxi}
		\country{China}}
	\email{ldx@stu.xidian.edu.cn}
	
	\author{Weiying Xie}
	\authornote{Corresponding author.}
	\orcid{0000-0001-8310-024X}
	\affiliation{%
		\institution{Xidian University}
		\city{Xi'an}
		\state{Shaanxi}
		\country{China}}
	\email{wyxie@xidian.edu.cn}
	
	\author{Mingxiang Cao}
	\orcid{0009-0000-1892-3645}
	\affiliation{%
		\institution{Xidian University}
		\city{Xi'an}
		\state{Shaanxi}
		\country{China}}
	\email{mingxiangcao@stu.xidian.edu.cn}
	
	\author{Yunke Wang}
	\orcid{0009-0003-9796-530X}
	\affiliation{%
		\institution{University of Sydney}
		\city{Sydney}
		\state{NSW}
		\country{Australia}}
	\email{yunke.wang@sydney.edu.au}
	
	\author{Yusi Zhang}
	\orcid{0009-0004-2179-2665}
	\affiliation{%
		\institution{Xidian University}
		\city{Xi'an}
		\state{Shaanxi}
		\country{China}}
	\email{23011210731@stu.xidian.edu.cn}
	
	\author{Leyuan Fang}
	\authornotemark[1]
	\orcid{0000-0003-2351-4461}
	\affiliation{%
		\institution{Hunan University}
		\city{Changsha}
		\state{Hunan}
		\country{China}}
	\email{leyuan\_fang@hnu.edu.cn}
	
	\author{Yunsong Li}
	\orcid{0000-0002-5239-6428}
	\affiliation{%
		\institution{Xidian University}
		\city{Xi'an}
		\state{Shaanxi}
		\country{China}}
	\email{ysli@mail.xidian.edu.cn}
	
	\author{Chang Xu}
	\orcid{0000-0002-4756-0609}
	\affiliation{%
		\institution{University of Sydney}
		\city{Sydney}
		\state{NSW}
		\country{Australia}}
	\email{c.xu@sydney.edu.au}
	
	\renewcommand{\shortauthors}{Daixun Li et al.}
	
	\begin{abstract}
		Multimodal image fusion and semantic segmentation are critical for autonomous driving.   Despite advancements, current models often struggle with segmenting densely packed elements due to a lack of comprehensive fusion features for guidance during training.   While the Segment Anything Model (SAM) allows precise control during fine-tuning through its flexible prompting encoder, its potential remains largely unexplored in the context of multimodal segmentation for natural images.
		In this paper, we introduce SAM into multimodal image segmentation for the first time, proposing a novel framework that combines Latent Space Token Generation (LSTG) and Fusion Mask Prompting (FMP) modules. This approach transforms the training methodology for multimodal segmentation from a traditional black-box approach to a controllable, prompt-based mechanism. Specifically, we obtain latent space features for both modalities through vector quantization and embed them into a cross-attention-based inter-domain fusion module to establish long-range dependencies between modalities. We then use these comprehensive fusion features as prompts to guide precise pixel-level segmentation. Extensive experiments on multiple public datasets demonstrate that our method significantly outperforms SAM and SAM2 in multimodal autonomous driving scenarios, achieving an average improvement of 4.1$\%$ over the state-of-the-art method in segmentation mIoU, and the performance is also optimized in other multi-modal visual scenes.
	\end{abstract}

	\ccsdesc[500]{Computing methodologies~Scene understanding}

	\keywords{Scene understanding, Computer vision}
	
	\begin{teaserfigure}
		\centering
		\includegraphics[width=1\textwidth]{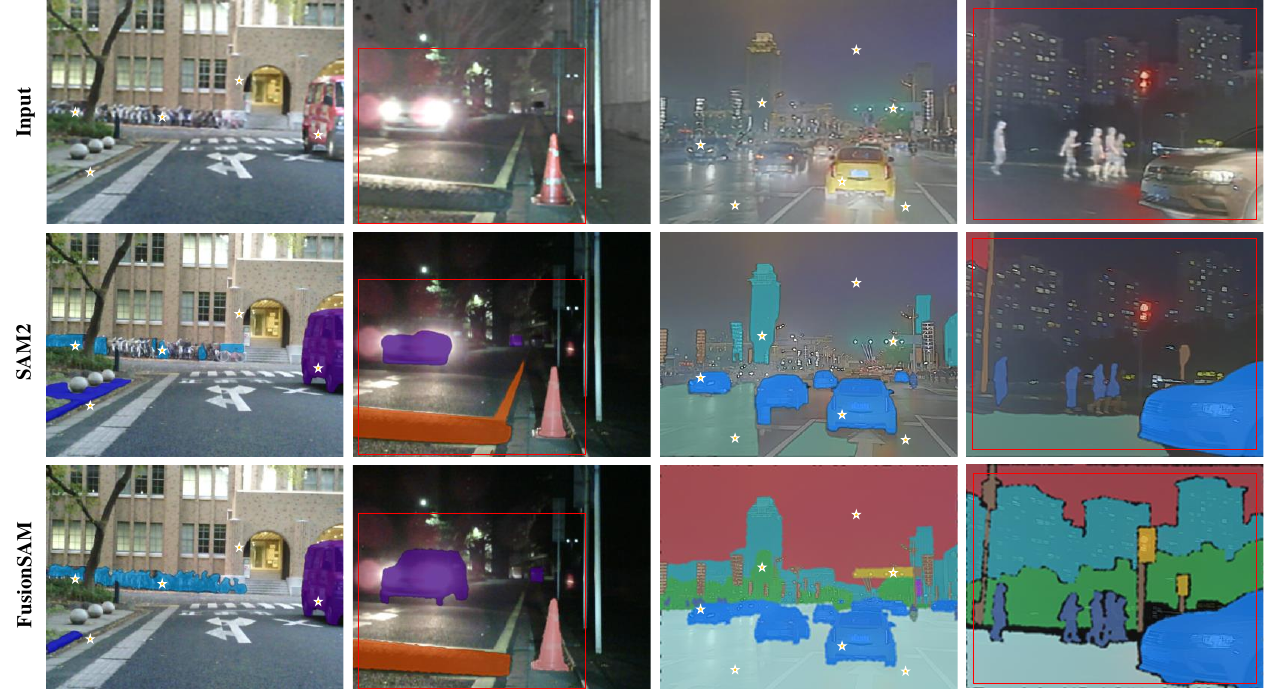}
		\caption{Comparative results of SAM2 and FusionSAM under the MFNet and FMB datasets: FusionSAM demonstrates superior boundary accuracy and structural completeness, while SAM2 struggles with misclassifications and unclear boundaries. To ensure fairness, the input is the fusion feature map of our method, annotations for points and boxes prompts are shown.}
		\label{fengmian}
		\vspace{2mm}
	\end{teaserfigure}

	\maketitle
	
	\newcommand\kddavailabilityurl{https://doi.org/10.5281/zenodo.15523136}
	
	\ifdefempty{\kddavailabilityurl}{}{
		\begingroup\small\noindent\raggedright\textbf{KDD Availability Link:}\\
		The source code of this paper has been made publicly available at \url{\kddavailabilityurl}.
		\endgroup
	}
	
	\section{Introduction}
	Accurate and comprehensive scene understanding is crucial for scene understanding and autonomous driving \cite{zhang2023visible, min2024driveworld, schon2021mgnet, liao2024vlm2scene, jiang2023vad}. Due to the limitations of sensor imaging devices, no single modality sensor can independently provide a complete description of the scene \cite{zhou2024general,xue2023dynamic,cao2023multi,xu2023murf}. For instance, infrared sensors capture thermal radiation information, highlighting objects of interest such as pedestrians and vehicles \cite{bellagente2024multifusion,sheinin2024projecting,erlenbusch2023thermal,jiang2024flexible,zhang2025irsam,zhao2024equivariant,yi2024text}. Conversely, visible light sensors capture reflected light, generating scenes rich in texture details \cite{liu2024coconet,pendota2024deep}. By combining these modalities, complementary details that might be missed by individual sensors can be captured, enhancing the model's ability to perform semantic segmentation of the complete scene \cite{cao2023autoencoder,mayr2024narrowing,li2024dstcfuse,zheng2024learning,ma2022swinfusion}. Therefore, the fusion of infrared and visible light images has become a mainstream solution for improving scene understanding and semantic segmentation. However, current semantic segmentation models struggle to comprehend densely packed elements in multimodal driving and other modality fusion scenes, failing to fully represent the captured information for better subsequent  segmentation results.
	
	In recent decades, advancements in semantic segmentation within deep learning have significantly propelled the understanding of multimodal scenes. Capturing efficient multimodal fusion representations is key to enhancing segmentation performance. A common approach involves feature-level fusion of infrared and visible light images using Convolutional Neural Networks (CNNs) to extract rich semantic representations, but the local  constraints of CNNs make it challenging to effectively merge information from different modalities.
	As an alternative, Transformer architectures, with their attention mechanisms and ability to model long-range dependencies, facilitate better global fusion and utilization of complementary information\cite{li2023efficient,zhang2024mrfs,peng2024transloc4d,rahman2024mist,yang2021focal}. 
	
	However, pure transformer architectures lack the flexibility required for scene understanding, especially in autonomous driving scenarios where elements are densely packed \cite{cao2023multimodal,huang2023video,ye2022rope3d}, and edge textures of segmented categories are blurred due to varying lighting conditions and nighttime environments. Without intermediate fine-tuning guidance to focus on critical regions, segmentation distortions can occur, hindering better scene parsing.
	The Segment Anything Model (SAM) has emerged as a transformative method for single-modal natural scene segmentation due to its flexible prompting architecture \cite{ravi2024sam,kirillov2023segment,wang2024sam}. Remarkably, the prompt architecture of SAM enhances the model's ability to focus on detailed features. Through the guiding mechanism of prompts, SAM can more effectively direct the segmentation process compared to transformers that lack fine-tuned control. This is crucial for the dense element segmentation required in autonomous driving scenarios. However, SAM has not yet been extensively studied in the realm of multimodal fusion.

	To address the challenges of traditional approaches that directly input linearly fused results into SAM, which often leads to excessive redundancy in the fused modalities and weakens the effective information from each modality, an important issue arises: how to efficiently fuse the modalities while ensuring the fusion features are as comprehensive as possible, and use these features as effective segmentation prompts for SAM. To overcome this, we innovatively propose FusionSAM, a Latent Space driven \textbf{S}egment \textbf{A}nything \textbf{M}odel for Multi-Modal Fusion and Segmentation, which endows SAM with efficient multimodal image fusion and segmentation capabilities. Specifically, we first capture latent space feature embeddings of the two modalities through vector quantization to obtain efficient downsampled representations. Then, we establish long-range dependencies between the modalities using a cross-attention-based inter-domain fusion module, capturing comprehensive information as fusion features to guide precise pixel-level segmentation. To the best of our knowledge, this is the first study to apply SAM to multimodal visual segmentation tasks in natural images, and it outperforms current state-of-the-art methods, as shown in Figure \ref{fengmian}. Our main contributions are as follows:

	\begin{itemize}
		\item[$\bullet$] We extend SAM to multimodal image segmentation in natural images for the first time. Through SAM's flexible prompt encoder we achieve efficient fusion and segmentation of multimodal images, meeting the complex requirements of autonomous driving scenarios with dense elements and varying lighting conditions.
		
		\item[$\bullet$] We propose a novel FusionSAM framework that includes the Latent Space Token Generation (LSTG) and Fusion Mask Prompting (FMP)  Module. By capturing latent space representations through vector quantization and performing cross-domain fusion of these features, we generate precise segmentation.
		
		\item[$\bullet$] Extensive experiments on public datasets and benchmarks show that FusionSAM significantly outperforms state-of-the-art methods, including SAM and SAM2, in multimodal autonomous driving  multi-modal visual scenes (RGB, Infrared, Depth) scenarios, achieving a notable 4.1\% improvement in segmentation IoU, validating its effectiveness and robustness. The performance is also optimized in other multi-modal visual scenes.
	\end{itemize}

	\begin{figure*}[htbp]
		\centering
		\includegraphics[width=1\linewidth]{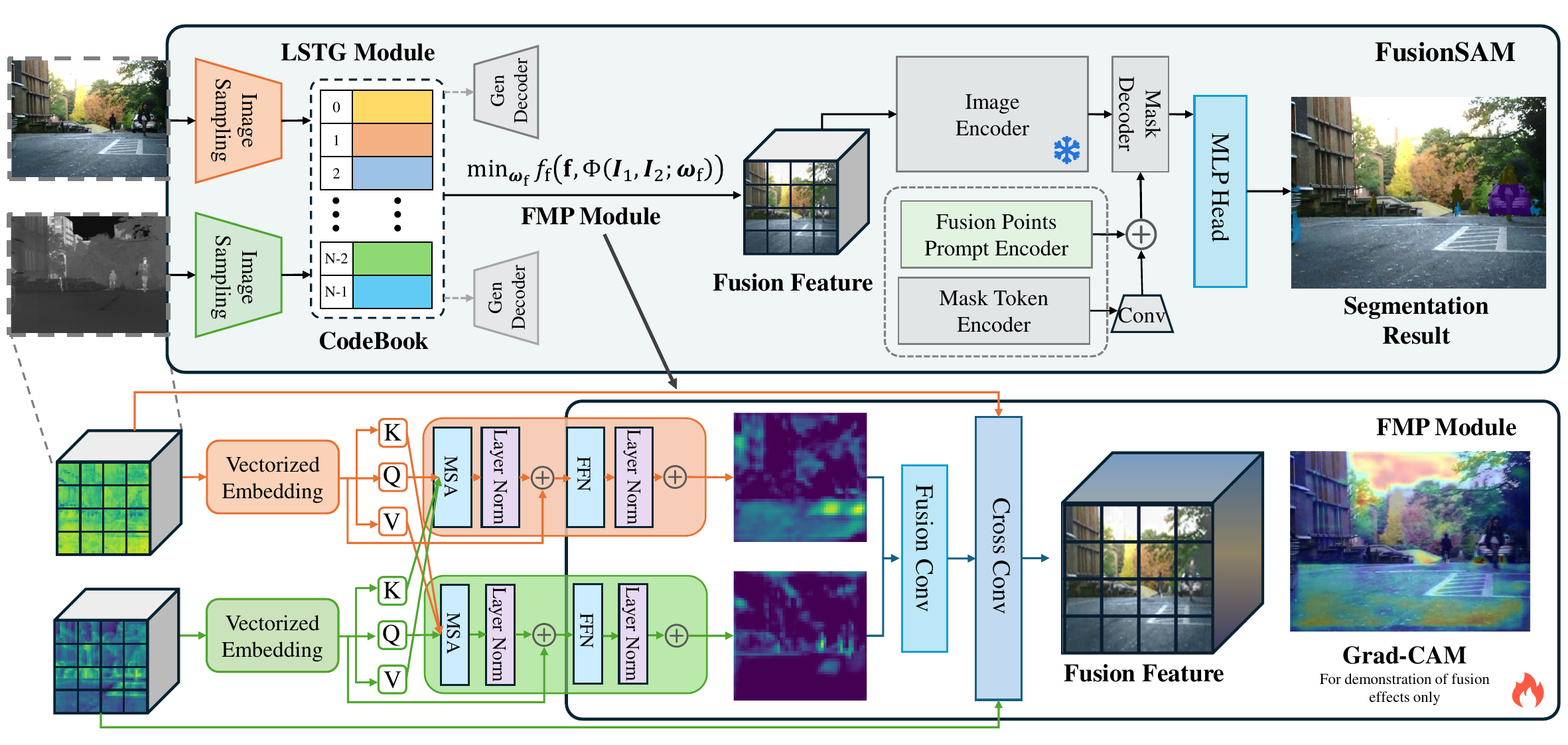}
		\centering
		\caption{Overview of \textbf{FusionSAM} framework for multimodal visual segmentation, which enhances multimodal visual understanding on the original SAM architecture. The main improvements include \textbf{Latent Space Token Generation (LSTG)} Module and \textbf{Fusion Mask Prompting (FMP)} Module. Unlike the traditional SAM architecture, the Fusion Point Prompt Encoder in FusionSAM extracts point prompts directly from the fusion features, enabling more comprehensive and precise guidance for segmentation. All parts of the architecture except the image encoder participate in the training phase.}
		\label{framework}
	\end{figure*}  
	
	\section{Related Work}
	\subsection{Segment Anything Model (SAM)}
	The SAM enables efficient object segmentation through simple prompt embeddings, like points or bounding boxes, guiding the model to focus on specific regions \cite{NEURIPS2023_5f828e38, Schon_2024_CVPR, Ren_2024_WACV, Shen_2024_CVPR, NEURIPS2023_1be3843e, ma2024segment, Huang_2024_CVPR,wei2024semantic,ren2024segment,li2023semantic}. Derived methods in single-modality segmentation include RobustSAM \cite{chen2024robustsam} by Chen \textit{et al.}, which improved SAM’s performance on low-quality images, and Crowd-SAM \cite{cai2024crowd} by Cai \textit{et al.}, which enhanced segmentation in crowded scenes with an Efficient Prompt Sampler and Part-Whole Discriminator Network. SAM has also been adapted for cross-modal tasks in fields like medical imaging and remote sensing. For example, Pandey \textit{et al.} used YOLOv8 and SAM for cross-modal segmentation \cite{pandey2023comprehensive}, while Yan \textit{et al.} introduced RingMo-SAM \cite{10315957} for segmenting optical and SAR data. However, these methods only linearly adapt SAM for multimodal tasks, missing the full potential of multimodal features. They also overlook SAM's powerful prompting architecture, which could better activate multimodal fusion features during training to guide segmentation. Our proposed FusionSAM, captures latent space representations through vector quantization, enabling comprehensive cross-domain fusion and using these features as precise segmentation prompts.

	\subsection{Multi-Modality Image Fusion}
	In autonomous driving, integrating various sensors is essential for accurate scene understanding, as single-modality data is insufficient \cite{Liang_2022_CVPR, li2024mdfl, zhang2021deep, 10.1145/3503161.3547902, NEURIPS2020_339a18de, Wang_2022_CVPR, Zhang_2020_CVPR_Workshops,zhang2024e2e,wang2024driving,wang2023learning,ando2023rangevit,li2023mseg3d}. Wang \textit{et al.} proposed AsymFusion \cite{wang2020learning}, which enhances multimodal feature interaction using a dual-branch structure with asymmetric fusion blocks. Zhang \textit{et al.} developed MRFS \cite{zhang2024mrfs}, combining CNN-based Interactive Gated Mixed Attention with transformer-based Progressive Cycle Attention to overcome bottlenecks in infrared-visible fusion. Feng \textit{et al.} introduced MAF-Net \cite{9864311}, which effectively segments road potholes by fusing RGB and disparity data. Ma \textit{et al.} proposed SwinFusion \cite{ma2022swinfusion}, leveraging cross-domain long-range learning and Swin Transformer for global information integration and complementary feature extraction. Most existing methods rely on convolutional networks or transformers, which struggle with global information extraction and flexible segmentation in dense scenes. To overcome these limitations, we apply multimodal fusion within SAM, using its flexible prompting to enhance segmentation in complex autonomous driving scenarios.

	\section{Methodology}

	\subsection{Problem Formulation}
	
	For the task of multimodal image fusion, we first assume visible image $I_1 \in \mathbb{R}^{H \times W \times C_{i n}}$ and infrared image $I_2 \in \mathbb{R}^{H \times W \times C_{i n}}$, where the two source images from different domains are aligned. Let $H$, $W$, and $C_{in}$ denote the height, width, and channel number of input images, respectively. As shown in the upper right corner of Figure \ref{framework}, to achieve pixel-level segmentation, we design an interactive neural network for fusion and segmentation, and optimize the model to find a set of optimal parameters. The optimization model is formulated as follows:

	\begin{equation}
		\min _{\boldsymbol{\omega}_{\mathrm{f}}, \boldsymbol{\omega}_{\mathrm{s}}} 
		f_{\mathrm{f}}\Big(I_f, \Phi\big(I_1, I_2 ; \boldsymbol{\omega}_{\mathrm{f}}\big)\Big)
		+ f_{\mathrm{s}}\Big(I_s, \Psi\big(I_1, I_2 ; \boldsymbol{\omega}_{\mathrm{s}}\big)\Big),
	\end{equation}
	$I_f \in \mathbb{R}^{H \times W \times C_{\mathrm{in}}}$ and $I_s \in \mathbb{R}^{H \times W \times C_{\mathrm{in}}}$ represent the fusion map and segmentation result, produced by the fusion network $\Phi$ and segmentation network $\Psi$ with learnable parameters $\boldsymbol{\omega}_{\mathrm{f}}$ and $\boldsymbol{\omega}_{\mathrm{s}}$. The functions $f_{\mathrm{f}}(\cdot)$ and $f_{\mathrm{s}}(\cdot)$ correspond to the objective functions for fusion and segmentation, measuring the discrepancies between the predictions and their respective targets.
	
	\subsection{Fusion Segment Anything Model}
	We propose FusionSAM, which enhances image fusion while preserving the segmentation capability of the SAM architecture. By integrating a fusion module that enables latent space representation embedding and cross-modal consistency fusion into the original SAM architecture, so that its performance will be greatly improved.

	\subsubsection{Model Overview}

	Figure \ref{framework} presents an overview of the proposed FusionSAM. The key contribution of FusionSAM is its Latent Space Token Generation (LSTG) and Fusion Mask Prompting (FMP) modules. Unlike methods that fine-tune or add adapters to SAM and SAM2, FusionSAM’s strength lies in its rigorous and well-considered approach to efficient multimodal fusion and segmentation. This efficiency is achieved by fusing compact and comprehensive latent space representations of both modalities, rather than the original large-scale images, enabling more thorough and effective fusion.
	
	\textbf{{Training.}}
	To train FusionSAM, we generate efficient fused modality representations, which are then input into model. Initially, a vector encoder creates latent space representations for both modalities, followed by cross-attention-guided fusion to achieve a comprehensive representation. Unlike the original SAM, we modify the input tokens for segmentation into Full-fledged Output Tokens (FOT), which are enhanced versions of the latent representations designed to capture the full spectrum of fused features for segmentation. These FOTs, along with the prompt token, are processed through the SAM decoding layers to generate the segmentation mask.
	
	The LSTG block processes the raw images from both modalities and transforms them into efficient latent space features. Simultaneously, the FMP module performs multimodal fusion on the obtained latent features. It uses cross-attention mechanisms to learn features from different modality domains, producing refined and comprehensive features. These refined fusion features are then fed into the mask encoder to enhance segmentation quality.
	
	In summary, the robust segmentation capability of the completed FusionSAM framework primarily stems from the training of the LSTG and the FMP modules. Additionally, the decoder and segmentation head from the original SAM architecture are also involved in the learning process. This integration ensures that the model comprehensively understands the fused features from both modalities, thereby enhancing segmentation performance.

	\textbf{{Inference.}}
	In the FusionSAM framework, the ViT-driven image encoder is not involved in training, it is solely used for inference to generate inputs for the mask decoder.

	\subsubsection{Latent Space Token Generation}
	In our multimodal image fusion and segmentation approach, the LSTG module effectively transforms complex input data from visible and infrared modalities into structured latent space representations. This transformation is essential for the efficient integration of diverse information sources. By drawing inspiration from Vector Quantized Generative Adversarial Networks (VQGAN) \cite{esser2021taming}, we enhance our model's capability to capture and fuse complementary features from both modalities, thereby improving the performance of multimodal tasks.

	Each image $I_i \in \mathbb{R}^{H \times W \times C}$ is transformed into a spatial set of codebook entries $I_i^q \in \mathbb{R}^{h \times w \times d_c}$, where $i \in \{1, 2\}$, $h = \frac{H}{s}$, $w = \frac{W}{s}$, $d_c$ is the latent dimensionality, and $s$ denotes the scaling factor. This transformation enables the efficient representation of complex multimodal features.
	
	The LSTG module employs an encoder $\mathcal{E}$ to compress the input images into latent vectors, capturing significant features necessary for multimodal integration:
	\begin{equation}
		z_i = \mathcal{E}(I_i) \in \mathbb{R}^{h \times w \times d_c}.
	\end{equation}
	
	These latent vectors preserve the critical multimodal characteristics needed for subsequent fusion and segmentation, allowing us to efficiently integrate and interpret complementary information from both the visible and infrared domains.
	
	The quantization process translates  encoder outputs \( z_i \) into discrete representations using a learned codebook \( \mathcal{C} \), aligning and structuring diverse features from both modalities for fusion:
	\begin{equation}
		I_i^q = \operatorname{Quant}(z_i) = \left( \operatorname{argmin}_{c_k \in \mathcal{C}} \| z_{ij} - c_k \| \right) \in \mathbb{R}^{h \times w \times d_c}.
	\end{equation}
	
	By mapping each latent vector $z_{ij}$ to the closest entry in the codebook, the $\operatorname{Quant}(\cdot) $ discretizes the latent representation, this function, aligns similar features from both modalities. This enhances the model's ability to merge complementary information and mitigate modality-specific noise.
	
	The decoder $\mathcal{G}$ reconstructs the original images from these quantized representations, ensuring that the fusion representation obtained in the subsequent fusion process retains the high fidelity and rich details required for accurate segmentation:
	\begin{equation}
		\hat{I}_i = \mathcal{G}(I_i^q) = \mathcal{G}(\operatorname{Quant}(\mathcal{E}(I_i))).
	\end{equation}
	
	To optimize the LSTG module for multimodal tasks, we incorporate a reconstruction loss $\mathcal{L}_{\text{rec}}$ to maintain the fidelity of each modality's essential features and a commitment loss $\mathcal{L}_{\text{commit}}$ to ensure effective codebook utilization:
	\begin{equation}
		\mathcal{L}_{\text{rec}} = \sum_i \| I_i - \hat{I}_i \|^2,
	\end{equation}
	\begin{equation}
		\mathcal{L}_{\text{commit}} = \sum_i \| \operatorname{sg}[z_i] - I_i^q \|^2_2 + \beta \| \operatorname{sg}[I_i^q] - z_i \|^2_2,
	\end{equation}
	where ${sg}[\cdot]$ denotes the stop-gradient operation. These loss functions help preserve crucial information while promoting the generalization capabilities of the model, which are vital for handling the complexities of multimodal data.
	
	To further enhance the representation quality, a perceptual loss $\mathcal{L}_{\text{perc}}$ and an adversarial loss $\mathcal{L}_{\text{adv}}$ are incorporated. These components focus on maintaining visual coherence and realism across the fused modalities:
	\begin{equation}
		\mathcal{L}_{\text{perc}} = \sum_i \| \Phi(I_i) - \Phi(\hat{I}_i) \|^2.
	\end{equation}
	\begin{equation}
		\mathcal{L}_{\text{adv}} = \sum_i \left( \log \mathcal{D}(I_i) + \log(1 - \mathcal{D}(\hat{I}_i)) \right),
	\end{equation}
	$\Phi (\cdot) $ represents the feature map of the reconstructed image in VQGAN, \(\mathcal{D}\) is the discriminator network used in the adversarial learning framework, distinguishing between real and generated data. These enhancements ensure that the model captures both low-level detail and high-level semantic information, which is crucial for effective multimodal segmentation.

	The LSTG module's ability to create a robust and structured representation from complex multimodal inputs is key to the successful integration and interpretation of diverse data sources. By minimizing redundancy while preserving critical information, these tokens facilitate seamless integration into our segmentation framework, significantly enhancing the model's capacity to discern and process complex scenes in multimodal environments. This ensures a comprehensive understanding and efficient handling of the diverse data inherent in visible and infrared images, making the LSTG module a vital component in our multimodal fusion strategy.
	
	As a result, we choose vector quantization in the Latent Space Token Generation (LSTG) module due to its ability to generate compact and semantically aligned latent representations across modalities.  Unlike raw features, VQ discretizes inputs into a codebook, making the fused features more structured and suitable for prompt extraction in the SAM architecture.  We found that this approach not only reduces spatial redundancy but also preserves modality-specific information in a consistent latent space, enabling more efficient and robust fusion. 
	
	\begin{figure*}[t]
		\centering
		\includegraphics[width=1\linewidth]{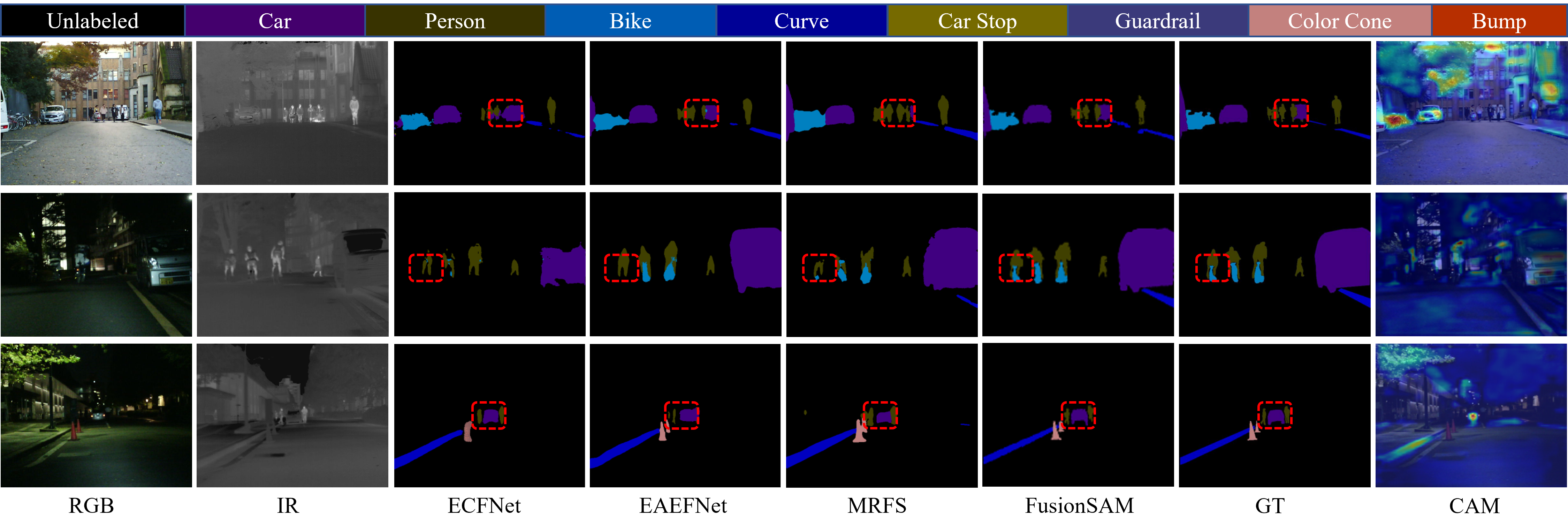}
		\centering
		\caption{A visual demonstration of different methods on the MFNet dataset.}
		\label{seg_mfnet}
	\end{figure*}

	\begin{table*}[!h]
		\centering
		\caption{Results of quantitative segmentation on the test set of MFNet dataset.}
		\setlength{\tabcolsep}{2.85mm}
		\begin{tabular}{c|ccccccccc|c}
			\hline
			Method & Unlabeled & Car & Person & Bike & Curve & Car Stop & Guardrail & Color Cone & Bump & mIoU(\%) \\ \hline
			EGFNet$_{23}$ & 97.7 & 87.6 & 69.8 & 58.8 & 42.8 & 33.8 & 7.3 & 48.3 & 47.1 & 54.8 \\ 
			SegMiF$_{23}$ & 98.1 & 87.8 & 71.4 & 63.2 & 47.5 & 31.1 & 0.0 & 48.9 & 50.3 & 56.1 \\ 
			EAEFNet$_{23}$ & 97.6 & 87.6 & 72.6 & 63.8 & 48.6 & 35.0 & 14.2 & 52.4 & 58.3 & 58.9 \\ 
			LASNet$_{23}$ & 97.4 & 84.2 & 67.1 & 56.9 & 41.1 & 39.6 & \textbf{18.9} & 48.8 & 40.1 & 54.9 \\ 
			SFAF-MA$_{23}$ & 97.0 & 88.1 & 73.0 & 61.3 & 45.6 & 29.5 & 5.5 & 45.7 & 53.8 & 55.5 \\ 
			ECFNet$_{24}$ & 98.0 & 85.7 & 73.5 & 59.7 & 45.7 & 36.7 & 4.0 & 47.4 & 55.1 & 56.2 \\ 
			MRFS$_{24}$ & 98.6 & 89.4 & {75.4} & 65.0 & 49.0 & 37.2 & 5.4 & \textbf{53.1} & 58.8 & 59.1 \\ 
			\rowcolor{gray!20}  \textbf{Ours} & \textbf{98.7} & \textbf{89.6} & \textbf{76.2} &\textbf{75.8}  & \textbf{69.6} & \textbf{50.1} & 10.2 & 47.6 & \textbf{61.4}& \textbf{64.4} \\ \hline
		\end{tabular}
		
		\label{mfnet}
	\end{table*}
	
	\subsubsection{Fusion Mask Prompting Module}
	
	As shown in Algorithnm \ref{alg:fmp}, this module integrates information from different domains presented by each modality into a unified fusion mask based on the SwinFusion. By leveraging the rich and comprehensive features present in the fusion representation as prompts, FMP module provides flexible fine-tuning guidance for the segmentation process, leading to improved segmentation performance. For instance, if multimodal fusion feature map contains complete information, using local area features as point prompts during training can further enhance the model’s segmentation accuracy.

	\begin{algorithm}[htbp]
		\caption{Fusion Mask Prompting (FMP) Module Fusion.}
		\begin{algorithmic}[1]
			\Require RGB feature map $\mathbf{F}_{\text{RGB}}$, Thermal feature map $\mathbf{F}_{\text{T}}$
			\Ensure Fused feature $\mathbf{F}_{\text{out}}$
			\State \textbf{Step 1: Self-Attention on Each Modality}
			\State $\mathbf{F}_{\text{RGB}}^{\text{self}} \gets \text{SelfAttention}(\mathbf{F}_{\text{RGB}})$
			\State $\mathbf{F}_{\text{T}}^{\text{self}} \gets \text{SelfAttention}(\mathbf{F}_{\text{T}})$
			
			\State \textbf{Step 2: Cross-Attention with Query from RGB and Key / Value from Thermal}
			\State $\mathbf{Q} \gets \mathbf{F}_{\text{RGB}}^{\text{self}} \cdot \mathbf{W}_Q$
			\State $\mathbf{K} \gets \mathbf{F}_{\text{T}}^{\text{self}} \cdot \mathbf{W}_K$
			\State $\mathbf{V} \gets \mathbf{F}_{\text{T}}^{\text{self}} \cdot \mathbf{W}_V$
			\State $\mathbf{A} \gets \text{softmax} \left( \frac{\mathbf{Q} \cdot \mathbf{K}^\top}{\sqrt{d_k}} \right)$
			\State $\mathbf{F}_{\text{fused}} \gets \mathbf{A} \cdot \mathbf{V}$
			
			\State \textbf{Step 3: Residual Connection and Layer Normalization}
			\State $\mathbf{F}_{\text{out}} \gets \text{LayerNorm}(\mathbf{F}_{\text{fused}} + \mathbf{F}_{\text{RGB}}^{\text{self}})$
			
			\Return $\mathbf{F}_{\text{out}}$
		\end{algorithmic}
		\label{alg:fmp}
	\end{algorithm}

	Specifically, FMP module begins with a cross-domain fusion unit that employs cross-attention mechanisms to establish long-range dependencies between different modality domains. This facilitates the exchange of \textbf{Queries} ($Q$), \textbf{Keys} ($K$) , and \textbf{Values} ($V$) across domains, ensuring the complete fusion of multimodal features. This process ensures that the fusion mask captures comprehensive interactions between the latent representations \( I_1^q \) and \( I_2^q \), enhancing the segmentation process by focusing on critical, contextually relevant features that are essential for understanding dense and complex scenes. The inter-domain mechanism is defined as:
	\begin{equation}
		\begin{aligned}
			& \left\{Q_1, K_1, V_1\right\} = \left\{I_1^q W_{Q1}, I_1^q W_{K1}, I_1^q W_{V1}\right\}, \\
			& \left\{Q_2, K_2, V_2\right\} = \left\{I_2^q W_{Q2}, I_2^q W_{K2}, I_2^q W_{V2}\right\}.
		\end{aligned}
		\label{fusion_1}
	\end{equation}
	\begin{equation}	
		\begin{aligned}
			& z_1' = \text{LN}\left(\text{softmax}\left(\frac{Q_{1} K_{2}^T}{\sqrt{d_k}}\right)V_{2}\right) + Q_1, \\
			& z_2' = \text{LN}\left(\text{softmax}\left(\frac{Q_{2} K_{1}^T}{\sqrt{d_k}}\right)V_{1}\right) + Q_2.
		\end{aligned}
		\label{fusion_2}
	\end{equation}
	
	Equations \ref{fusion_1} and \ref{fusion_2} present the process of the \textbf{Fusion Conv} of Figure \ref{framework}.
	LN($\cdot$) is the layer normalization, whcih always performed after feed forward network, the outputs \( z_1' \) and \( z_2' \) represent the globally fused features, which are then processed through a convolutional layer, generating a fused representation \( z_c \) that encapsulates the essential information from both modalities. This fused representation serves as the initial fusion mask, guiding the segmentation by highlighting the regions of interest identified through the cross-domain fusion process.

	To further enhance the fusion mask, the FMP module integrates a  complementary feature fusion unit, which emphasizes the unique characteristics of each modality while ensuring the complete integration of global features. This unit introduces a complementary feature fusion mechanism, where the two modalities are first fused through a cross-attention mechanism to produce \( z_0\) , which encapsulates the distinctive features of each individual modality. This result is then combined with the initial fusion mask \( z_0\), strengthening the segmentation prompt by leveraging the comprehensive information from both approaches:

	\begin{equation}
		\begin{aligned}
			& \left\{Q_0, K_0, V_0\right\} = \left\{z_0^q W_{Q1}, z_0^q W_{K1}, z_0^q W_{V1}\right\}, \\
			& \left\{Q_f, K_f, V_f\right\} = \left\{z_f^q W_{Q2}, z_f^q W_{K2}, z_f^q W_{V2}\right\}.
		\end{aligned}
		\label{cross1}
	\end{equation}
	\begin{equation}
		\begin{aligned}
			I_f &= \text{LN}\left( \text{softmax}\left(\frac{Q_f K_o^T}{\sqrt{d_k}}\right)V_o \right) + z_c .
		\end{aligned}
		\label{cross2}
	\end{equation}

	Equations \ref{cross1} and \ref{cross2} present the process of the \textbf{Cross Convolutional layer} of Figure \ref{framework}. \( z_f^q \) represents the result of directly performing \textbf{Fusion Convolutional layer} on the original latent space embeddings of the two modalities.
	The final representation  is then processed through a convolutional layer to produce the  fusion mask $I_f$, which serves as a precise prompt for guiding pixel-level segmentation. In previous SAM models, prompts were typically embedded as masks in the form of points or bounding boxes to guide segmentation. However, in FusionSAM, we innovate by using fused multimodal images as prompts. The "fusion mask" refers to this integrated representation of both modalities, allowing the model to leverage comprehensive feature information for more accurate segmentation. This approach enhances the model’s ability to handle complex multimodal environments.

	By leveraging these cross-domain and complementary feature fusion units, the FMP effectively captures comprehensive fusion features that are critical for accurate segmentation. The integration of global context and long-range dependencies ensures that the model can differentiate between foreground and background elements, even in densely packed autonomous driving scenes. This comprehensive approach allows SAM to achieve robust and high-fidelity segmentation results, effectively addressing the challenges of multimodal image fusion.

	The final fused representation $I_f$, derived from the FMP, is fed into the original SAM framework's image encoder. This encoder processes the multimodal fusion results, transforming them into high-dimensional features that encapsulate the rich information from the visible and infrared modalities. The encoded features are then input into the mask decoder, which utilizes a modified transformer architecture to generate mask features through a series of attention operations. Finally, the decoder's output, representing the refined segmentation, is further processed by a multilayer perceptron (MLP) classification head, ensuring that the model accurately identifies and distinguishes between distinct regions within the input data.
	
	\begin{figure*}[t]
		\centering
		\includegraphics[width=1\linewidth]{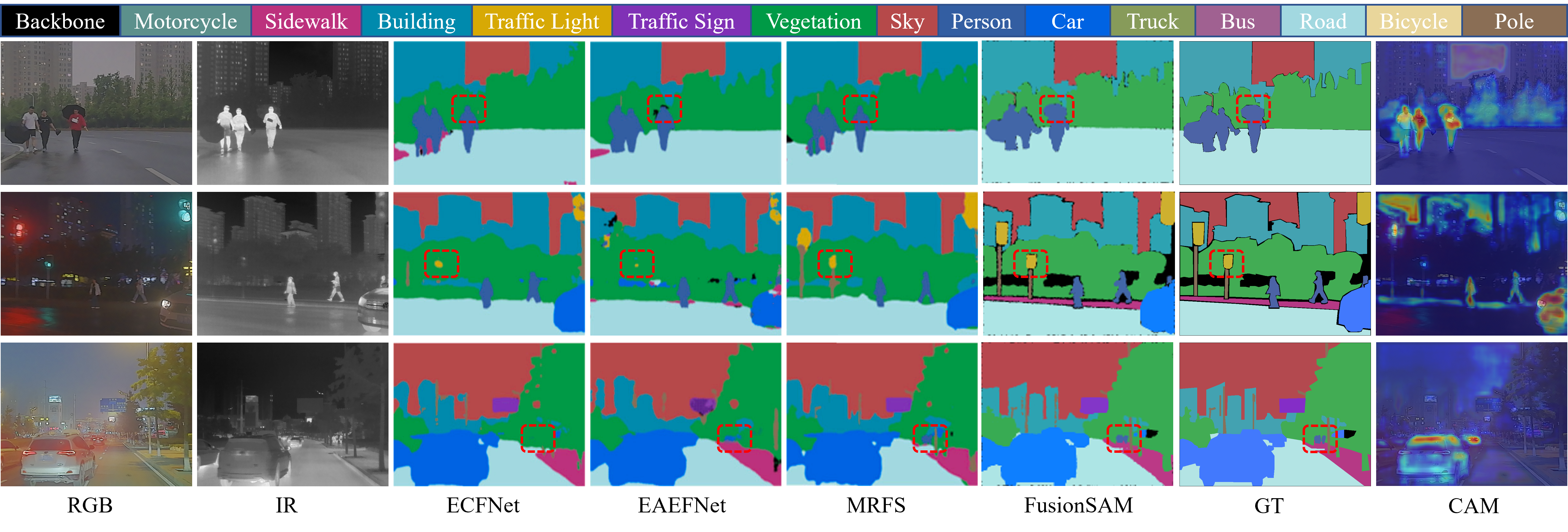}
		\centering
		\caption{A visual demonstration of different methods on the FMB dataset.}
		\label{seg_fmb}
	\end{figure*} 
	\begin{table*}[h]
		\centering
		\caption{Results of quantitative segmentation on the test set of FMB dataset.}
		\setlength{\tabcolsep}{4mm}
		\begin{tabular}{c|cccccccc|c}
			\hline
			Method & Car & Person & Truck & T-Lamp & T-Sign & Building & Vegetation & Pole & mIoU(\%) \\ \hline
			EGFNet$_{23}$ & 77.4 & 63.0 & 17.1 & 25.2 & 66.6 & 77.2 & 83.5 & 41.5 & 47.3 \\ 
			SegMiF$_{23}$ & 78.7 & 65.5 & 42.4 & 35.6 & 71.7 & 80.1 & 85.1 & 35.7 & 58.5 \\  
			EAEFNet$_{23}$ & 79.7 & 61.6 & 22.5 & 34.3 & 74.6 & 82.3 & 86.6 & 46.2 & 58.0 \\  
			LASNet$_{23}$ & 73.2 & 58.3 & 33.1 & 32.6 & 68.5 & 80.8 & 83.4 & 41.0 & 55.7 \\  
			SFAF-MA$_{23}$ & 73.0 & 55.7 & 14.3 & 13.6 & 54.2 & 73.0 & 78.9 & 38.1 & 42.7 \\  
			ECFNet$_{24}$ & 80.0 & 63.1 & 12.8 & 40.6 & 71.9 & 81.4 & 84.4 & 44.6 & 52.5 \\  
			MRFS$_{24}$ & 76.2 & \textbf{71.3} & 34.4 & \textbf{50.1} & \textbf{75.8} & 85.4 & 87.0 & \textbf{53.6} & 61.2 \\  
			\rowcolor{gray!20} \textbf{Ours} & \textbf{80.4} & 59.4 & \textbf{48.2} & 47.3 & 52.1 & \textbf{85.7} & \textbf{87.8} & 52.0 & \textbf{64.1} \\ \hline
		\end{tabular}
		\label{fmb}
	\end{table*}

	\section{Experiment}
	\subsection{Experimental Setups}
	\textbf{Datasets.}
	Two representative datasets, including MFNet \cite{ha2017mfnet} and FMB \cite{liu2023multi}, containing 1569 and 1500 pairs of visible and infrared images with resolutions of 480$\times$640 and 600$\times$800, respectively, to train and evaluate our method. Annotated into 9 and 14 categories relevant to autonomous driving and semantic understanding, these datasets offer varied lighting conditions and rich scenes that enhance the generalization ability of fusion and segmentation models. 
	In addition, we select the VDT-2048 dataset, which consists of 2,048 pairs of RGB, thermal (T), and depth (D) images, spanning three distinct modalities. This dataset provides a comprehensive collection of multimodal data, designed to test and evaluate the performance of segmentation models across varying environmental conditions.
	
	\begin{figure*}[t]
		\centering
		\includegraphics[width=1\linewidth]{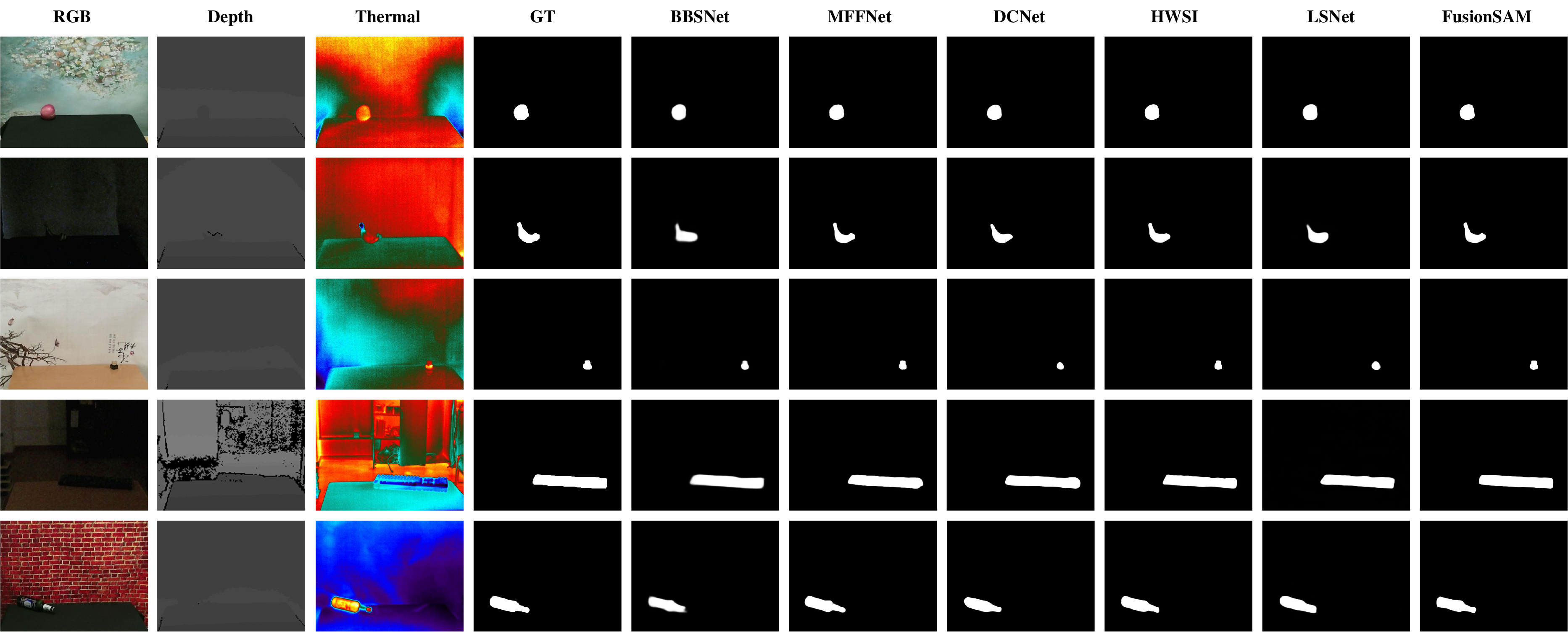}
		\centering
		\caption{Qualitative demonstrations of different approaches on the VDT-2048 dataset. FusionSAM displays an RGB+D+T visualization. Our method shows optimal results in terms of texture details.}
		\label{new_exp}
	\end{figure*}

	\begin{table*}[t]
		\centering
		\caption{Comparison results of $S, F_\beta^{\max }, F_\beta^{\text {mean }}, F_\beta^{a d p}, E_{\xi}^{\max }, E_{\xi}^{\text {mean }}, E_{\xi}^{a d p}$ and $M A E$ on VDT-2048 for RGB-T SOD task.   " $\uparrow$ " (" $\downarrow$ ") means that the larger (smaller) the better. The best three results in each row are marked in red, green,The table contains a variety of visual modes including RGB+Depth, RGB+Thermal, and the contrast accuracy of RGB+Depth+Thermal. }
		\setlength{\tabcolsep}{2.2mm}
		\scalebox{1.15}{
			\begin{tabular}{c|c|cccccccc}
				\hline
				Methods     & Modality & $S$      & $MAE$    & $E_{\xi}^{adp}$ & $E_{\xi}^{mean}$ & $E_{\xi}^{max}$ & $F_{\beta}^{adp}$ & $F_{\beta}^{mean}$ & $F_{\beta}^{max}$  \\
				\hline
				BBSNet$_{21}$\cite{zhai2021bifurcated}     & RGB+D      & 0.9117 & 0.0046 & 0.8747 & 0.9357  & 0.9769 & 0.6957 & 0.8267  & 0.8679     \\
				\rowcolor{gray!20} \textbf{Ours} & \textbf{RGB+D}    & \textbf{0.9407} & \textbf{0.0036} & \textbf{0.63} & \textbf{0.9512}  & \textbf{0.9841} & \textbf{0.8707} & \textbf{0.8767}  & \textbf{0.8912} 	  \\
				\hline
				DCNet$_{22}$\cite{tu2022weakly}  & RGB+T      & 0.8787 & 0.0038 & 0.9658 & 0.9436  & 0.9659 & 0.8521 & 0.8450  & 0.8525       \\
				LSNet$_{23}$\cite{zhou2023lsnet}      & RGB+T    & 0.8867 & 0.0044 & 0.9327 & 0.9631  & 0.9737 & 0.7607 & 0.8097  & 0.8354     \\
				\rowcolor{gray!20} \textbf{Ours} & \textbf{RGB+T}    & \textbf{0.9496} & \textbf{0.0029} & \textbf{0.9705} & \textbf{0.9790}  & \textbf{0.9850} & \textbf{0.8822} & \textbf{0.8947}  & \textbf{0.9056} 	  \\
				\hline
				HWSI$_{22}$\cite{song2022novel}        & RGB+D+T    & 0.9318 & 0.0026 & 0.9815 & 0.9845  & 0.9890 & 0.8718 & 0.8961  & 0.9102     \\
				MFFNet$_{23}$\cite{wan2023mffnet}      & RGB+D+T    & 0.9394 & 0.0025 & 0.9831 & 0.9825  & 0.9905 & 0.8758 & 0.9034  & 0.9180     \\
				
				\rowcolor{gray!20} \textbf{Ours} & \textbf{RGB+D+T}    & \textbf{0.9542} & \textbf{0.0020} & \textbf{0.9890} & \textbf{0.9875}  & \textbf{0.9952} & \textbf{0.8982} & \textbf{0.9108}  & \textbf{0.9266} 	  \\
				\hline
			\end{tabular}
		}	
		\label{new_table}
	\end{table*}
	
	\textbf{Training Details.}
	During 100 epochs of training, multimodal images are subjected to 4$\times$ downsampled features by the LSTG module, and the FMP module further captures efficient fusion representations. Our initial learning rate is set to 1e-4, using the Adam optimizer with a weight decay of 1e-3, the batch size is set to 4, and vit/h is used as the encoder. All experiments are performed on a NVIDIA A100 Tensor Core GPU.
	We use mean Intersection over Union (mIoU) to evaluate semantic segmentation performance. mIoU is the average of intersection over the union for each class.

	\subsection{Comparison with SAM}
	SAM \cite{kirillov2023segment} is competitive in the segmentation field because of its powerful segmentation performance and adaptability in different fields. Compared with SAM, SAM2 \cite{ravi2024sam} has significant improvements in applicable fields, segmentation accuracy, and running speed. To demonstrate the effective design and powerful performance of our FusionSAM and maintain a fair comparison, we use SAM and SAM2 to directly infer the fused feature maps generated in FusionSAM, and the results are shown in Figure \ref{xiaorongbiao}\textcolor{red}{.(a)}. The SAM series cannot handle multimodal image segmentation, whereas our method introduces SAM into the multimodal field, ensuring its excellent segmentation performance and expanding its applicability in more complex scenarios.

	\subsection{Comparison with State-of-the-Arts}
	We conduct comparative experiments and evaluations with seven state-of-the-art semantic segmentation methods, including EGFNet \cite{zhou2022edge}, SegMiF \cite{liu2023multi}, EAEFNet \cite{liang2023explicit}, LASNet \cite{li2022rgb}, SFAF-MA \cite{he2023sfaf}, ECFNet \cite{shen2024ecfnet}, and MRFS \cite{zhang2024mrfs}. We provide quantitative results in Tables \ref{mfnet} and \ref{fmb}. Our FusionSAM achieves the highest mIoU on both datasets. Compared with the second-highest method, FusionSAM improves mIoU by 5.3\% and 2.9\% on MFNet and FMB, respectively. More specifically, for heat-insensitive categories, such as Bike, Car Stop, Building, Curve, and Truck, our method achieves significant superiority due to the effective visual quality preservation and enhancement. Overall, these findings confirm that our method achieves SOTA excellence in semantic segmentation.
	
	As shown in Figure \ref{new_exp} and Table \ref{new_table}, we further demonstrate the generalization capability of FusionSAM by evaluating it on the RGB+Depth and RGB+Thermal modalities, as well as the combined RGB+Depth+Thermal modality. Our experiments show that FusionSAM consistently outperforms existing SOTA methods across all modalities. The fusion of all three modalities is performed using cross-attention to model the long-range dependencies between the features, which allows for the effective integration of complementary information from the different modalities. These results further highlight the robustness and effectiveness of our method across multiple input configurations, reinforcing its superiority in multimodal semantic segmentation tasks.

	\subsection{Ablation Study}
	\begin{figure}
		\centering
		\includegraphics[width=\linewidth]{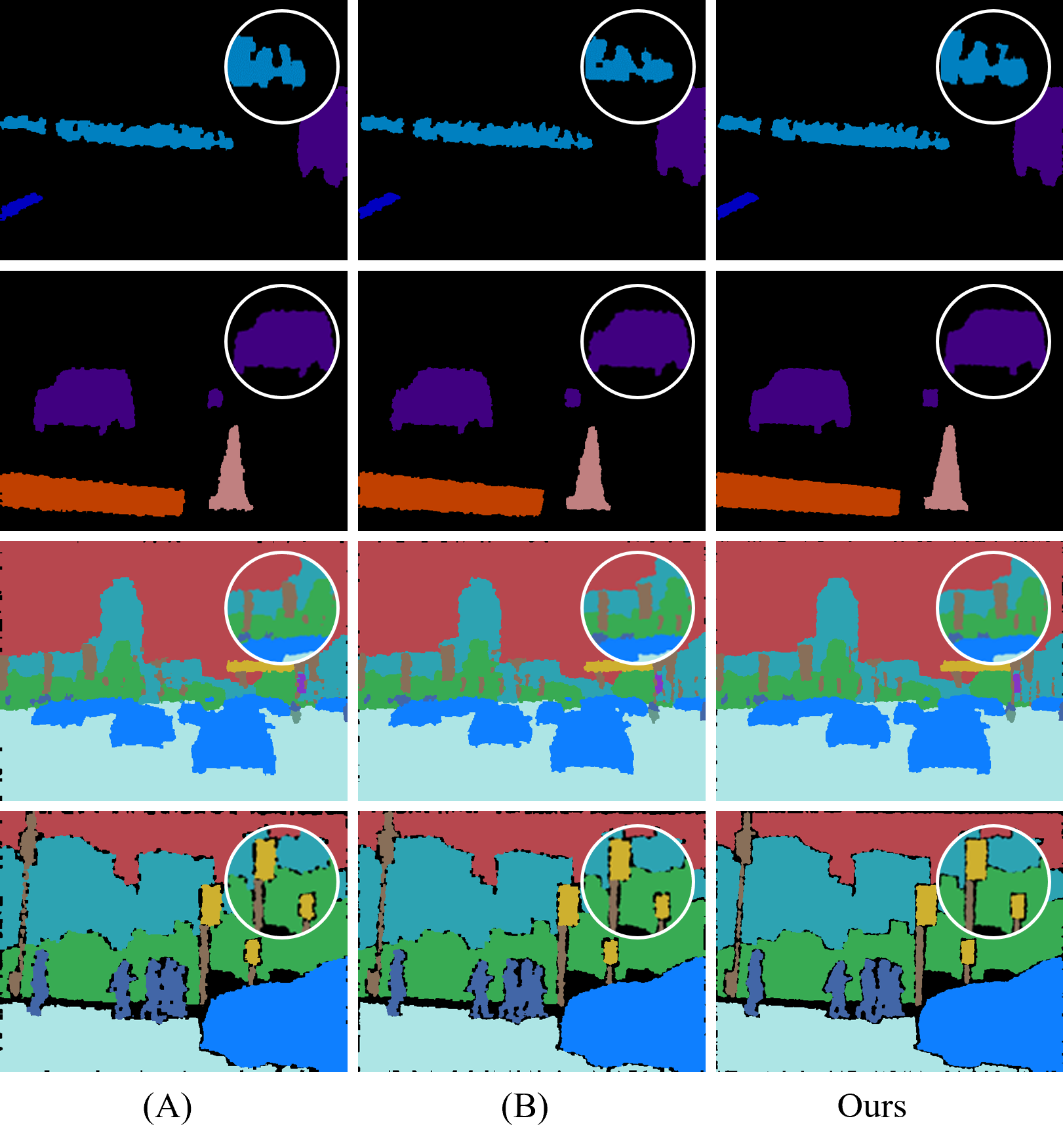}
		\vspace{-8mm}
		\caption{Visualization of ablation studies in FusionSAM.}
		\label{xiaorong}
	\end{figure}
	
	\begin{figure}[!t]
		\centering
		\includegraphics[width=\linewidth]{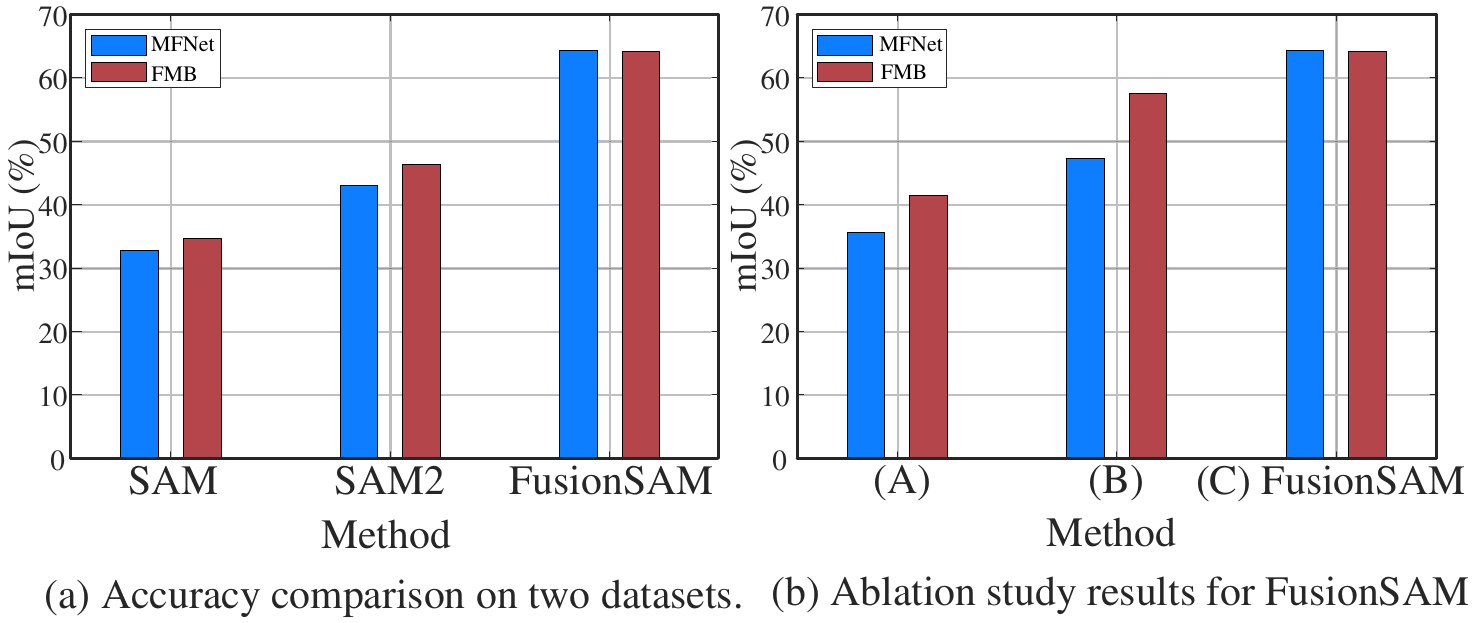}
		\centering
		\caption{(a) Comparison of FusionSAM's performance with the existing baseline models SAM and SAM2; (b) Performance comparison of FusionSAM in ablation experiments.}
		\label{xiaorongbiao}
	\end{figure}
	
	To explore the contribution of each part of our method in detail, we designed three scenarios: (A) Omitting the LSTG module compared to our FusionSAM; (B) Removing the FMP module from the fusion process and replacing it with direct concat; (C) Complete FusionSAM. The results of the ablation experiment are shown in Figure \ref{xiaorong}. We can observe that FusionSAM achieves the best segmentation results on both datasets. As shown in Figure \ref{xiaorongbiao}\textcolor{red}{.(b)}, in (A) , by removing the LSTG module, we notice that the results drop by 28.8\% and 22.7\%, respectively, while resulting in poor segmentation results, which shows the effectiveness of the LSTG module in generating latent space tokens through vector quantization. Our fusion method is verified in (B). Without introducing the fusion mask hint, the model has difficulty distinguishing the foreground and background, ignoring the unique and complementary features of each modality, resulting in a decrease in mIoU of 17.1\% and 6.5\%, respectively. Therefore, our proposed LSTG and FMP module can effectively improve the segmentation performance of multimodal images and produce excellent visual results.
	
	\subsection{Result Visualization}

	Figures \ref{seg_mfnet} and \ref{seg_fmb} show segmentation visualizations and Class Activation Mapping (CAM) of our method on the MFNet and FMB datasets, and compare with the most competitive methods. These datasets present segmentation challenges due to their rich categories, complex imaging conditions, and diverse scene details. Existing fusion methods struggle to highlight dim infrared targets (e.g., bicycles in Figure \ref{seg_mfnet}, second row) and recognize distant pedestrians (Figure \ref{seg_fmb}, third row). Methods relying on two-stream networks often introduce conflicts if feature fusion is incomplete, leading to misclassifications, such as occluded cars (Figure \ref{seg_mfnet}, first row) and human shapes (Figure \ref{seg_fmb}, first row). Additionally, edge blurring in dense target predictions is common (Figure \ref{seg_mfnet}, third row). By embedding latent space representations and achieving cross-modal consistency, our method reduces redundancy while retaining key information, significantly improving SAM's segmentation performance and enabling accurate object classification across diverse scenes.

	\section{Conclusion}
	
	A key challenge in multimodal semantic segmentation for autonomous driving is efficiently fusing multimodal data as prompts to guide high-performance segmentation in dense scenes. Traditional methods often lead to redundancy and weakened modality information. To address this, we propose FusionSAM, a latent space-driven SAM framework that enhances multimodal fusion and segmentation. By utilizing vector quantization, we achieve non-redundant, efficient fusion while preserving modality-specific information. This is the first study to apply SAM to multimodal segmentation of natural scenes with fusion as a guiding prompt. Extensive experiments demonstrate that FusionSAM significantly outperforms state-of-the-art methods across multimodal visual scenes (RGB, infrared, depth), showing robustness in varying lighting conditions and offering a novel solution for future tasks.
	
	\begin{acks}
		This work was supported in part by National Natural Science Foundation of China under Grant 62425109, 62322117, 62371365, U24B20136, and U22B2014, and the Fundamental Research Funds for the Central Universities under Grant ZYTS24101. This work was supported by the Fundamental Research Funds for the Central Universities, and supported by the Innovation Fund of Xidian University under Grant YJSJ25007.
		This work was supported in part by the Australian Research Council under Projects DP240101848 and FT230100549.
	\end{acks}
\balance
\bibliographystyle{ACM-Reference-Format}
\bibliography{sample-base}


\end{document}